\documentclass[final]{cvpr}

\usepackage{times}
\usepackage{epsfig}
\usepackage{graphicx}
\usepackage{amsmath}
\usepackage{amssymb}
\usepackage{mathtools}
\usepackage{caption}
\usepackage{subcaption}
\usepackage{enumitem}

\usepackage{pifont}%
\newcommand{\cmark}{\ding{51}}%
\usepackage{multirow}

\usepackage[pagebackref=true,breaklinks=true,colorlinks,bookmarks=false]{hyperref}

\def\cvprPaperID{1377} %
\def\confYear{CVPR 2021}
\usepackage{booktabs}
\usepackage{cuted}
\usepackage{capt-of}

\usepackage{capt-of,etoolbox}	
\makeatletter
\apptocmd\@maketitle{{\myfigure{}\par}}{}{}
\makeatother
\begin{document}

\title{View Generalization for Single Image Textured 3D Models}
\author{Anand Bhattad\textsuperscript{1}\thanks{work done while interning with NVIDIA} \and Aysegul Dundar\textsuperscript{2,3} \and Guilin Liu\textsuperscript{3} \and Andrew Tao\textsuperscript{3} \and Bryan Catanzaro\textsuperscript{3} \and
\textsuperscript{1}University of Illinois at Urbana-Champaign \and \textsuperscript{2}Bilkent University \and \textsuperscript{3}NVIDIA}
\newcommand\myfigure{%
\centering
\vspace*{-0.156in}
\includegraphics[width=\textwidth]{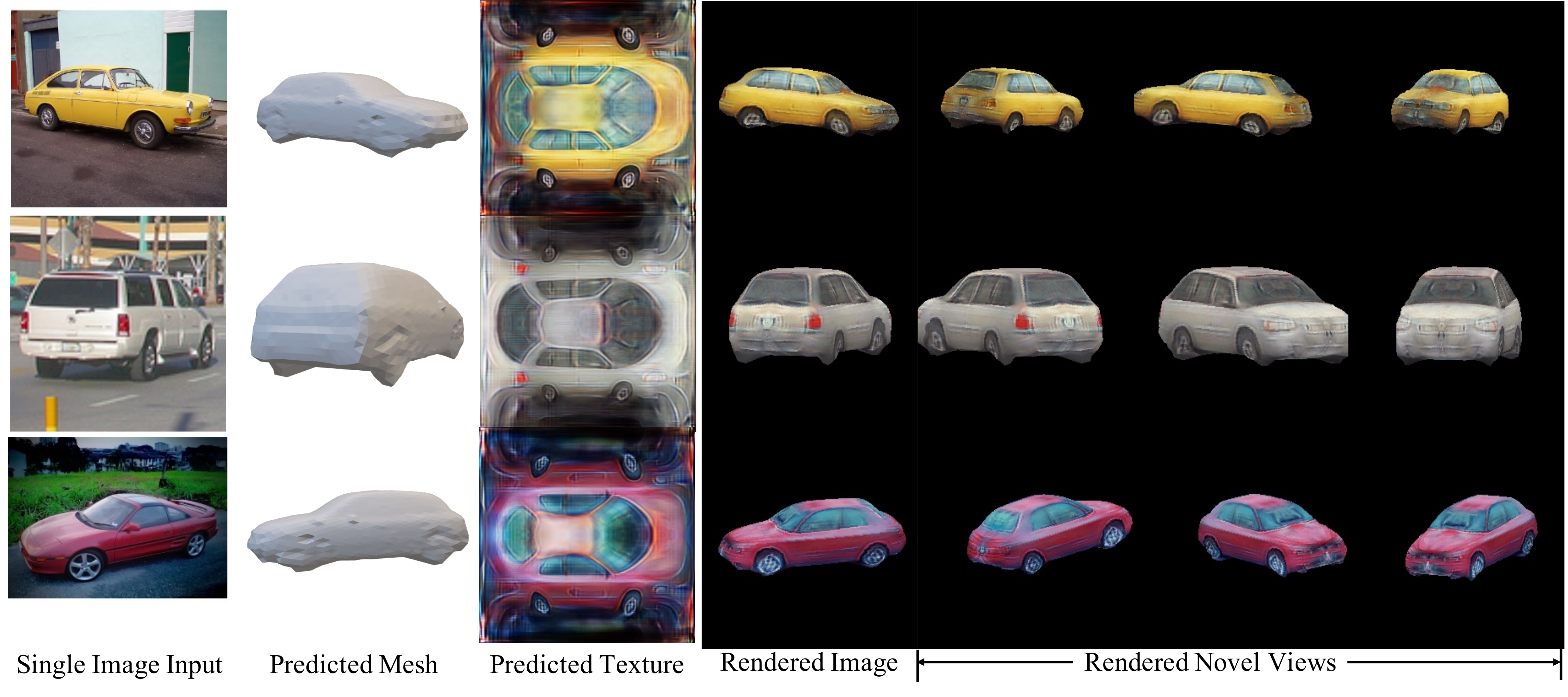}
\\[-0.13in]
\captionof{figure}{Given a single 2D image input, we infer high quality textured 3D models – a triangular mesh and a texture map,
which can then be used to render novel views. Our inferred models compare well with the source image when rendered from
that view.  More important, our method displays good {\em view generalization} -- new views of an
inferred model look like real pictures of that object. Project page: \href{https://nv-adlr.github.io/view-generalization}{https://nv-adlr.github.io/view-generalization}}\label{fig:teaser}
~\\[2mm]
}
\maketitle

\newcommand{\fix}{\marginpar{FIX}}
\newcommand{\new}{\marginpar{NEW}}
\begin{abstract}
Humans can easily infer the underlying 3D geometry and texture of an object only from a single
2D image.  Current computer vision methods can do this, too, but suffer
from {\em view generalization} problems --
the models inferred tend to make poor predictions of appearance
in novel views.  As for generalization problems in machine learning,
the difficulty is balancing single-view accuracy (cf. training error; bias) with
novel view accuracy (cf. test error; variance).  We describe a class of models whose
geometric rigidity is easily controlled to manage this tradeoff.  We
describe a cycle consistency loss that improves view generalization
(roughly, a model from a generated view should predict the original view
well).  View generalization of textures requires that models share
texture information, so a car seen from the back still has headlights
because other cars have headlights.  We describe a cycle consistency
loss that encourages model textures to be aligned,
so as to encourage sharing. We compare our method against the state-of-the-art method and show both
qualitative and quantitative improvements.

\vspace{-10pt}
\end{abstract}
\section{Introduction}
\begin{figure*}[!ht]
  \centering
  \includegraphics[width=0.97\linewidth]{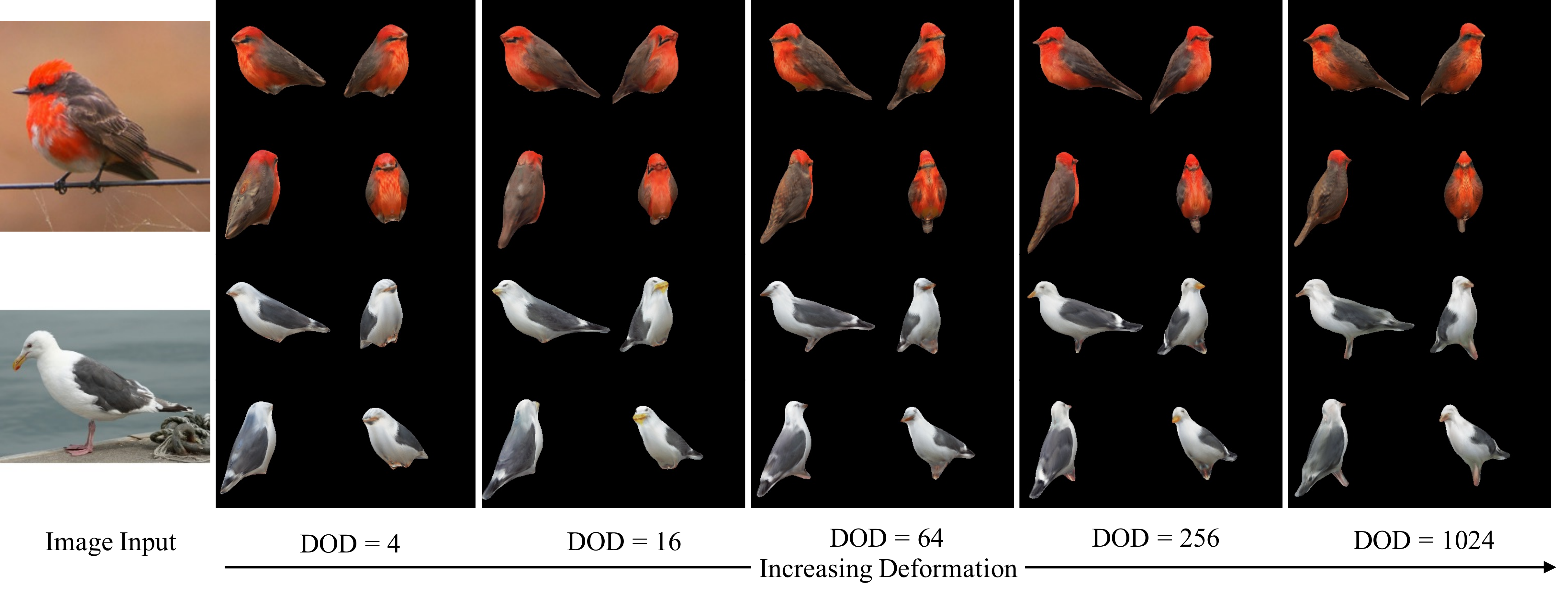}
  \vspace{-13pt}
  \caption{{\bf Controllable Deformation.} 
 Our approach is built upon a 2D convolutional deformation where one can easy control
the extent of deformation by changing degrees of deformation (DOD) or spatial resolution of our predicted 2D deformation
map.   A higher DOD generally yields a more flexible model (eg the beak of the gull), with better fit to the source view but
the possibility of view generalization problems (eg gull's eye in 1024). We report results for a fixed DOD, but note that specific instances might
perform better at different DOD's -- an artist might choose the best DOD for a particular source image.
  }
  \vspace{-15pt}
  \label{fig:degree_of_deformation}
 \end{figure*}
People can easily infer the 3D geometry and texture of an object from
a single 2D image.   Likely, we can reason
about the shape and the appearance of an object seen from
only a single view because we have seen many objects from many views.
Experience enables us to infer a reasonable estimate of shape and 
appearance of a novel object from a familiar category, even though
there is within category variation. Current algorithms can recover
3D shape and texture from a single image, but suffer from severe
\emph{view generalization} problems:  the recovered representation, viewed in a new camera, tends to look unrealistic. 
Either the inferred 3D mesh is squashed when viewed from other viewpoints  \cite{kanazawa2018learning, chen2019learning} or the inferred texture are muddled \cite{kanazawa2018learning}, and also not sharp enough with missing details \cite{chen2019learning}. For a faithful recovery of 3D models, current state-of-the-art (SOTA) methods rely on a finely curated multi-view datasets or use of synthetic images \cite{chang2015shapenet, choy20163d} which do not translate well to real image domains.
Therefore, in this work, we focus on inferring realistic textured 3D models -- meshes and texture maps; from a collection of single-view real images and by using only available 2D supervision. We not only synthesize superior results from the original image view but also our synthesis generalizes across views with a coherent geometry and convincing texture prediction (Fig.\ref{fig:teaser}).

Good view generalization requires a geometric deformation model that
can balance bias (source view) and variance (novel views).  A highly flexible model may fit the original
image well, but look bad from other directions; too rigid a model may generalize
moderately well, but not fit the original image.   Furthermore, good view
generalization requires texture models that can share details across instances
as appropriate: the way to know that this car has (say) headlights, even though they
are not visible in the original view, is to know that all cars have headlights, and borrow
headlights from some other model.  

Our geometric model uses a controllable convolutional deformation
approach that predicts a single UV deformation map over a
spherical surface.   This allows us to control the extent of deformation by varying the
predicted spatial map resolution. The convolutional deformation is previously proposed by \cite{pavllo2020convolutional}, here we use it for controllability by changing the degrees of deformation 3( Fig.~\ref{fig:degree_of_deformation}).
Our texture model is, again, on a single UV
map, allowing us to use cycle consistency losses to share texture features.
Our geometry and texture models are controlled with
two consistency losses: i) Rotation GAN cycle consistency
to deal with deformation from canonical views, encouraging
their appearance and shape to be realistic as the original
view images. ii) Texture mesh alignment consistency loss
that shares appearance information across images providing
strong spatial cues for inferring occluded texture.

Our experiments show that our framework built upon these philosophies and principles strongly improves the overall fidelity of synthesized textured 3D models (Fig.~\ref{fig:teaser}).
We show that our framework allows us to control the flexibility of our deformation that is specifically useful for deforming non-rigid object categories like birds where different images undergo different degrees of deformation (Fig.~\ref{fig:degree_of_deformation}). We also show that leveraging a few handfuls of multiple mesh templates instead of a single mesh template can improve overall recovered 3D inference quality without requiring the need of finely curated multi-view dataset.

Current SOTA methods are evaluated on intersection over union (IoU)
compared to the ground truth mask from the original view.
But a good mIoU from the original view may generate meshes that look poor from other views. The evaluation misses the point of 3D models - they are intended
to predict appearance from {\em new} viewpoints.   
We evaluate mIoU for comparison, but evaluate view generalization by
comparing multiple renderings from novel viewpoints with
real images of real objects (using Fr\'{e}chet
Inception Distance (FID) scores  \cite{heusel2017gans}).  We also conduct user studies comparing our
inferred 3D models with the baseline methods.
In summary, our main contributions are:
\begin{itemize}[leftmargin=*]
  \setlength\itemsep{-0.3em}
  \vspace{-5pt}
    \item A controllable convolutional deformation approach for better recovery of 3D geometry of the objects. 
    \item Two novel cycle consistency losses that strongly improves overall inferred textured 3D models.
    \item High-fidelity textured 3D model synthesis from a single image both qualitatively and quantitatively.%
\end{itemize}

\section{Related Work}

\textbf{Generative Adversarial Networks (GANs).}
Recently GAN~\cite{goodfellow2014generative} based models have achieved realistic image synthesis on various objects~\cite{karras2019style, zhang2019self, karras2020analyzing, yu2021dual}.
However, for this technology to be used in gaming, simulators, and virtual applications, they need to be fully controllable.
Recent work controls image generation by conditioning output on different type of inputs such as natural and synthetic images ~\cite{karacan2016learning, zhu2017unpaired, liu2017unsupervised, Zhao2018layout, dundar2020panoptic, mardani2020neural}, landmarks~\cite{lorenz2019unsupervised, wang2019few, dundar2020unsupervised}, and semantic maps~\cite{isola2017image, wang2018high, park2019semantic}.
Among these methods,~\cite{lorenz2019unsupervised,dundar2020unsupervised, jeon2020cross} disentangle images into pose and appearance in an unsupervised way, and show control over pose during inference. 
However, since they do not disentangle images into their 3D attributes, the control and manipulation over generated images are still limited.
Other models~\cite{karras2019style, zhang2019self, karras2020analyzing} are shown to learn an implicit 3D knowledge of objects without 3D supervision and the viewpoint of the synthesized object can be manipulated by latent codes.
InfoGAN~\cite{chen2016infogan} and $\beta$-VAE~\cite{higgins2016beta} are
proposed to learn interpretable disentangled features in an unsupervised manner.
However, there is no clear factorization between 3D object models and object appearance.
HoloGAN~\cite{nguyen2019hologan} for more explicit disentanglement, additionally incorporates a 3D rigid-body transformation
module and a 3D-to-2D projection module in the network, and learns 3D voxel feature representations of the world.
It provides the ability to control the pose of the generated objects better, however,  still lacks explicit physical 3D model.

\textbf{Single Image Textured 3D Inference.} Differentiable renderers~\cite{loper2014opendr, kato2018neural, liu2019soft, chen2019learning, ravi2020accelerating} enable inferring 3D properties from 2D images via gradient based optimization~\cite{kanazawa2018learning, goel2020shape, li2020self, henderson2020leveraging}. 
However, inferring 3D shape, texture, camera parameters, and lighting from single 2D images is inherently ill-posed.
To circumvent this problem, multi-view image datasets have been used especially on synthetic datasets~\cite{chang2015shapenet, tulsiani2017multi, tulsiani2018multi}. Multi-view images are not available for many datasets, and learning multi-view consistent shape and texture predictions from single image datasets is an important research area.
In that spirit,~\cite{pavllo2020convolutional} is similar to our work where GAN training is utilized to discriminate invisible and visible patches from mesh and texture predictions, enforcing the network to generate consistent  predictions among visible and invisible areas. 
In~\cite{pavllo2020convolutional}, a two step training pipeline is proposed; first texture and mesh networks are trained via a reconstruction loss to predict visible regions and second these networks are trained from scratch with GAN loss such that invisible and visible areas look unrecognizable.
The final networks do not encode 3D attributes from images but instead generate them from Gaussian noise which makes it not comparable to ours.

\begin{figure*}[!ht]
  \centering
  \includegraphics[width=\linewidth]{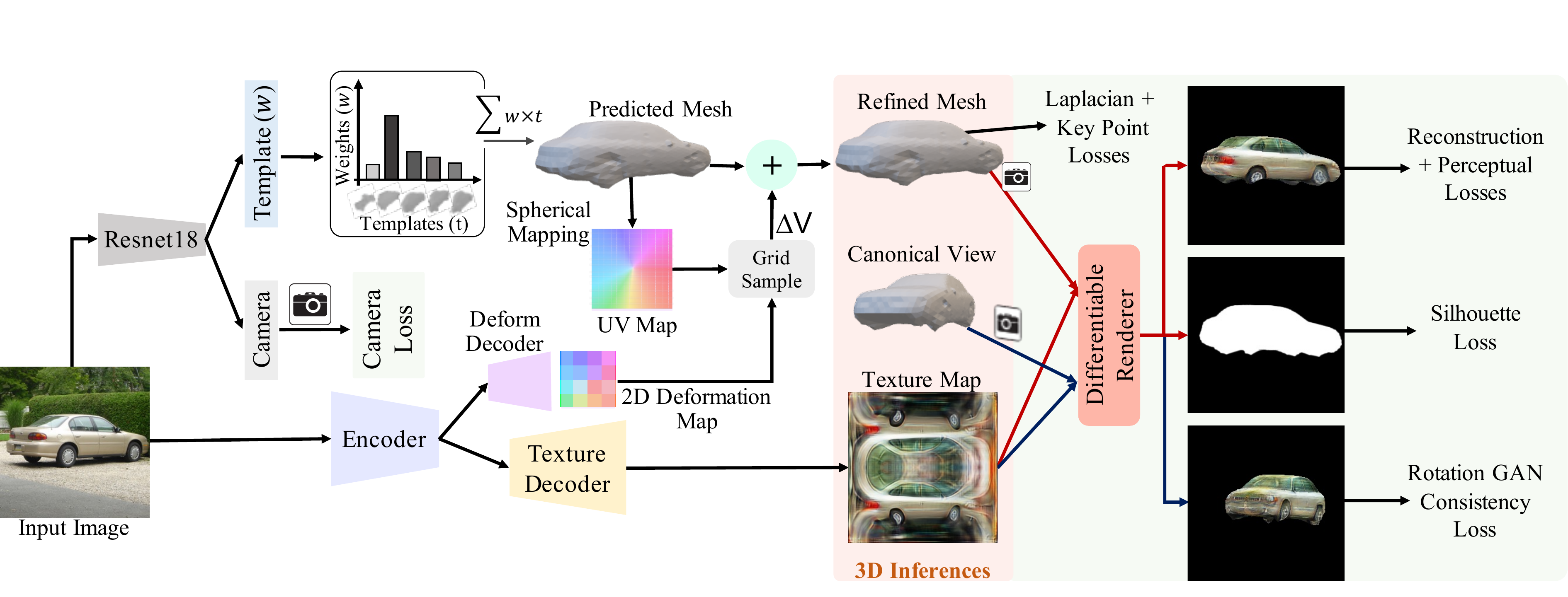}
    \vspace{-24pt}
  \caption{{\bf Architecture overview.}
  From a single input image, we learn to infer object's 3D geometry (a mesh) and its texture (an UV map). Our final geometry is learned by a weighted combination of object templates and refined by applying adequate deformation to it. For the deformation, we follow a convolutional approach over a spherical representation mapped as an UV map and sample final deformations of our predicted mesh template from this deformation map. We also learn camera parameters with provided ground-truth, but during training we render the prediction with a known camera for the stability. We use various losses that are based on 2D; an image loss between input image and the rendered prediction, a silhouette loss between the object mask and  prediction, and various other regularization and consistency losses described in Section~\ref{sec:app}.}
  \label{fig:over_method}
  \vspace{-10pt}
 \end{figure*}
\textbf{Cycle Consistency.} Cycle Consistency enforces that transforms along the cycles are close to the identity function. It has been used on various computer vision tasks from image translation~\cite{zhu2017unpaired, bansal2018recycle}, video interpolation~\cite{reda2019unsupervised, liu2019deep} to optical flow estimation~\cite{meister2018unflow} just to name a few.
Similar to our work, cycle consistency has been extensively used to disentangle image attributes such as pose and appearance disentanglement for person re-identification problem~\cite{zhou2020unsupervised, zheng2019joint}, for cross identity motion transfer for arbitrary objects~\cite{jeon2020cross}, shape and pose disentanglement for 3D meshes~\cite{zhou2020unsupervised}, and learn textures from 2.5D skecteches~\cite{zhu2018visual}.
A rotation consistency~\cite{zhou2020rotate} loss has been used to train pix2pixHD~\cite{wang2018high}, a 2D image-to-image translation method for rotating faces.
That work takes advantage of 3D face modeling by using off-the shelf 3D-fitting network~\cite{zhu2017face}, and renders images in novel views with missing textures. Pix2pixHD is trained to inpaint missing parts in the images.
In our work, we use cycle consistency among rotations of the same object and texture mesh alignment between two different objects when training encoders for 3D geomerty and texture predictions.

 \section{Approach}
 \label{sec:app}
 
 First, we start with introducing the overall pipeline as given in  Fig.~\ref{fig:over_method}, and then introduce the cycle consistency losses which provide a strong supervision signal for parts that are not visible from the original view.
 
 \subsection{Controllable Convolutional Deformation}
 We follow a similar encoder-decoder architecture from Chen $\etal$~\cite{chen2019learning} that encodes images into 3D geometry and decodes textures, and then both are fed into a differentiable renderer to reconstruct input images with image-view camera parameters as shown in Fig.~\ref{fig:over_method}. We use DIB-R~\cite{chen2019learning} as our differentiable renderer. 
 
 The key to infer strong 3D geometry is a detailed local control of model complexity and mesh deformation. This is crucial because we want the model to be flexible enough for every image and rigid enough for realistic rendering of novel views. Previous methods~\cite{kanazawa2018learning, chen2019learning} use a fully connected linear layer to predict deformation of a template mesh. This strategy has several disadvantages. First, each vertex is independently deformed. Second, there is no controllability over deformation when using a linear layer; that is if they need to be large (flexible) or small (rigid). Specially, for non-rigid objects like birds, one needs to be sufficiently flexible about the extent of deformation. Finally, these deformations are bound to a specific set of vertices of a template mesh and are difficult to re-sample for any new vertices (say learn deformation for low-poly mesh but synthesis a high-poly mesh at inference).

 In contrast to previous methods, we use a spherical mapping to map our surface, as represented by a mesh to a sphere~\cite{pavllo2020convolutional}. Then our deformation is a representation on the function of sphere directly with a fixed surface topology. We predict a single spatial deformation UV map for this spherical representation using a convolutional architecture as shown in  Fig.~\ref{fig:over_method}. We then sample deformation for each of the template's vertex position from the predicted UV deformation map.
 We show that this deformation allows us to better control deformation complexity as can be seen from Fig.~\ref{fig:degree_of_deformation}.
 For large deformation models, we predict a high-resolution spatial deformation map ($32\times32$) allowing models to be flexible enough. For small deformation, we predict a small-resolution deformation map ($2\times2$ or $4\times4$) having a stronger constraint over deformation as we deform only limited positions of the sphere.
 We refer these controllability as degree of deformation (DOD). For eg., $DOD = 4$ means our 2D deformation is of $2\times2$ spatial resolution. Since, the learned deformation is a function of sphere and not the template we could then also easily sample deformation for as many vertices without crippling learning and computational efficiency. We leverage this and infer 3D meshes with over $2500$ vertices and $6000$ faces and are about $4\times$ more dense than previous methods.
 
{\noindent \bf Mesh templates}. In our overall architecture, we follow a different approach than previous methods when learning geometry. We take advantage of provided 3D mesh templates from datasets if available.
Previous methods~\cite{kanazawa2018learning, chen2019learning} start with mean templates and learn to deform it.
We instead use multiple templates provided by PASCAL3D+ dataset~\cite{xiang2014beyond}, and a mean template for CUB dataset~\cite{welinder2010caltech}.  The network learns w, the template weight, to choose one from $n$ templates it needs to start with for an image. However, we do not have any supervision for these templates, so we predict normalized weights ($w$) of each template ($t$) and take their weighted sum as our predicted mesh template. We also use a learnable scale-factor ($s$) for each of these templates to adapt our templates to appropriate size. We then refine this predicted mesh by sampling deformation of all vertices ($\Delta V$) using the convolutional UV deformation map.
Our final refined mesh vertices are $V = \sum_{i=1}^{n}s_i\times w_i\times V^{t}_i + \Delta V$.

We show that making use of only a few additional templates improves our textured 3D inference significantly without requiring multi-view dataset.  For pre-processing of mesh templates, we use same procedure described in~\cite{kulkarni2019csm}. We use all meshes that are provided in PASCAL3D+ dataset and ignore those templates that fails during the following preprocessing steps. We first remove all interior triangles and faces by simplifying mesh into voxels. Next, we convert these voxels into a mesh by running a marching cubes algorithm. We then further simplify mesh by closing holes and decimating mesh. Finally, we use spherical parameterization and map sphere to our simplified mesh.

{\noindent \bf{Losses.}} Our base losses are same as DIB-R~\cite{chen2019learning} and use DIB-R as our strong baseline. We use $\mathcal{L}_1$ image reconstruction loss between the input image ($I$) and the rendered image ($I_r$) and also perceptual losses from Alexnet ($\Phi$) at different feature layers ($j$) between these images. 
\vspace{-2pt}
\begin{equation}
\label{eq:recon}
\small
    \mathcal{L}_{recon} = ||I - I_r ||_{1}
\end{equation}
\begin{equation}
\label{eq:percep}
\small
    \mathcal{L}_{percp} = ||\Phi_{j}(I) - \Phi_j(I_r) ||_2
\end{equation}
For shapes, we use an IoU loss between the silhouette rendered ($S_r$) and the silhouette ($S$) of the input object.
\vspace{-1pt}
\begin{equation}
\label{eq:sil}
\vspace{-1pt}
\small
\mathcal{L}_{sil} = 1 - \frac{||S\odot S_r ||_1}{||S+S_r - S\odot S_r ||} 
\end{equation}
Similar to~\cite{chen2019learning, liu2019soft}, we also regularize predicted mesh using \\
a smoothness loss and laplacian loss ($\mathcal{L}_{lap}$) constraining neighboring mesh triangles to have similar normals. Chen et al~\cite{chen2019learning} uses camera as latent by not learning them and using the ground truth cameras. Similar to~\cite{kanazawa2018learning}, we learn to predict camera pose 
using a simple $\mathcal{L}_2$ regression loss ($\mathcal{L}_{cam}$) between ground truth and predictions that we get from Resnet18 backbone as shown in our Fig.~\ref{fig:over_method}. We also predict keypoints from the inferred mesh and apply $\mathcal{L}_2$ regression loss comparing to the ground truth key point annotations ($\mathcal{L}_{KP}$). Finally, we regularize deformation to be small and hence minimize overall deformation to be as close to zero ($\mathcal{L}_{deform} = ||\Delta V||$). Following are our base losses:
\vspace{-2pt}
\begin{equation}
\begin{multlined}
\small
\mathcal{L}_{baseline} = 
\lambda_r \mathcal{L}_{recon} + \lambda_p \mathcal{L}_{percp} + \lambda_s \mathcal{L}_{sil} \\  + \lambda_c \mathcal{L}_{cam} + \lambda_{kp} \mathcal{L}_{KP} + \lambda_d \mathcal{L}_{deform} + \lambda_{lap} \mathcal{L}_{lap}   
\end{multlined}
\end{equation}

Our $\lambda$ coefficients are: $\lambda_r=20$, $\lambda_p=0.5$, $\lambda_s=5.0$,  $\lambda_c=1$, $\lambda_{kp}=50.0$, $\lambda_d=2.5$, $\lambda_{lap}=5.0$.

\subsection{Cycle-Consistency Losses}
We want our learning problem to be set in such a way that both texture and geometry information is shared across image collections efficiently while training and hence they can be inferred accurately during run-time.
Our use of multiple templates largely improves over inferring geometry without extreme deformation of mesh triangles for views other than what that can be observed.
We also benefit significantly by introducing two novel cycle-consistency losses such that both texture and geometry information are coherently shared across image collections. 

\begin{figure}[t]
  \centering
  \includegraphics[width=\linewidth]{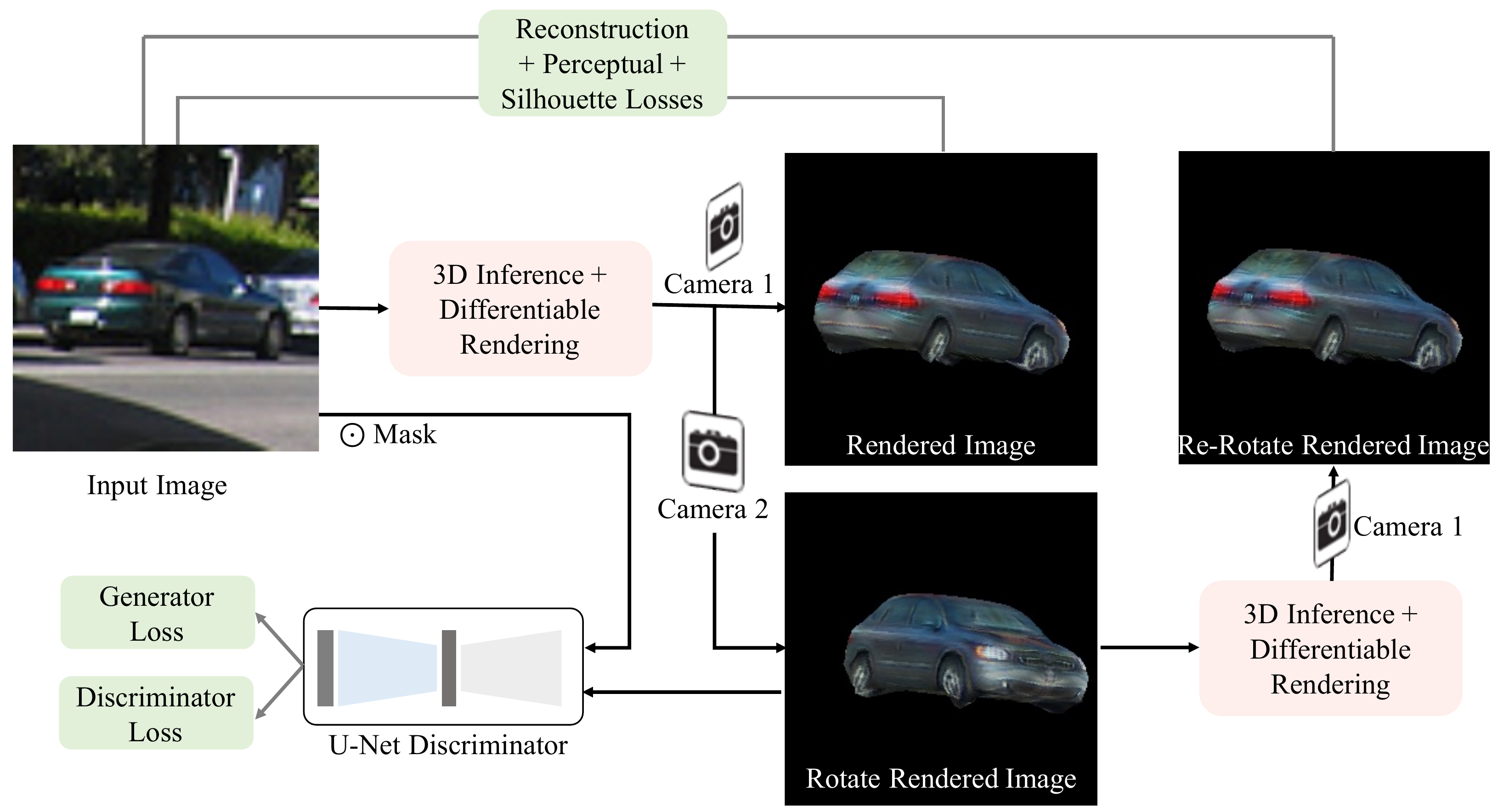}
  \vspace{-22pt}
  \caption{{\bf Rotation GAN Cycle Consistency}.  We extract geometry and texture predictions from input images, and render them from novel views. From those rendered images, we again predict geometry and texture but this time render them with the original camera parameters. The final images should match the input images. The rendered novel views must also appear realistic to a discriminator and is also trained with a generative loss.
  \vspace{-12pt}}
  \label{fig:rot_gan}
 \end{figure}

Our first loss is a GAN based rotation cycle consistency loss.  We use GAN losses on rendered images from novels views to be synthesized as real along with implicitly imposing a multi-view cycle consistency  constraint (Fig.~\ref{fig:rot_gan}). Our second loss is a texture-mesh alignment cycle consistency loss. We use this to align texture to meshes irrespective of their views and shapes (Fig.~\ref{fig:text_cyc}) and share information whenever possible. For example, the predicted headlights in texture map from one car image should be near about the same place when predicted from another car image.   
\vspace{-3pt}
\subsubsection{Rotation GAN Cycle Consistency} 
Single image 3D inference is an ill-posed problem. There is no supervision from other views and hence gradients to update geometry and texture.~\cite{wang2018pixel2mesh, liu2019soft} control these effects by imposing a multi-view consistency loss and renders multiple views given a single image on synthetic generated Shapenet dataset~\cite{chang2015shapenet}. 
However, this approach is not scalable for real images as it is hard to collect multi-view images. 

We impose multi-view consistency implicitly without requiring image-collections from multiple views.
The cycle based multi-view consistency is imposed as follows: say $\text M$ is a function which outputs 3D meshes given an input image, $\text T$ is a texture network, and $\text R$ is a differentiable renderer function that generates the image. 
Given an image $X_1$ and its corresponding camera pose $C_1$, we first infer this image into its geometric and texture representations, and render an intermediate image from a novel camera view, $C_2$.
 \vspace{-5pt}
 \begin{equation*}
   I_1 =  \text R(\text M(X_1), \text T(X_1), C_2)
      \vspace{-5pt}
  \end{equation*}
Using this intermediate image, we infer back the geometry and texture representations, and render back an image with the original view camera pose, $C_1$.
  \vspace{-5pt}
 \begin{equation*}
      X_1^{'}  =  \text R(\text M(I_1), \text T(I_1), C_1)
 \vspace{-5pt}
 \end{equation*}
We apply a cycle consistency check between $X_1$ and $X_1^{'}$ using the reconstruction loss (Eq~\ref{eq:recon}) and perceptual loss (Eq~\ref{eq:percep}).
Additionally, we train a U-Net discriminator~\cite{schonfeld2020u} that can distinguish if the image is real or fake at global level as well as individual pixels level.
For fake images, we use the novel view images, $I_1$, which provides gradient signals from the intermediate image.
These losses implicitly impose multi-view consistency; GAN loss penalizes if rendered novel views are not realistic for both the shape and appearance, and the cycle consistency loss regularizes the network to infer predictions for occluded regions consistent with the original view. This constraint significantly improves our overall 3D synthesis. 
\vspace{-3pt}
\subsubsection{Texture Mesh Alignment Cycle Consistency}
\begin{figure*}
  \centering
  \includegraphics[width=\linewidth]{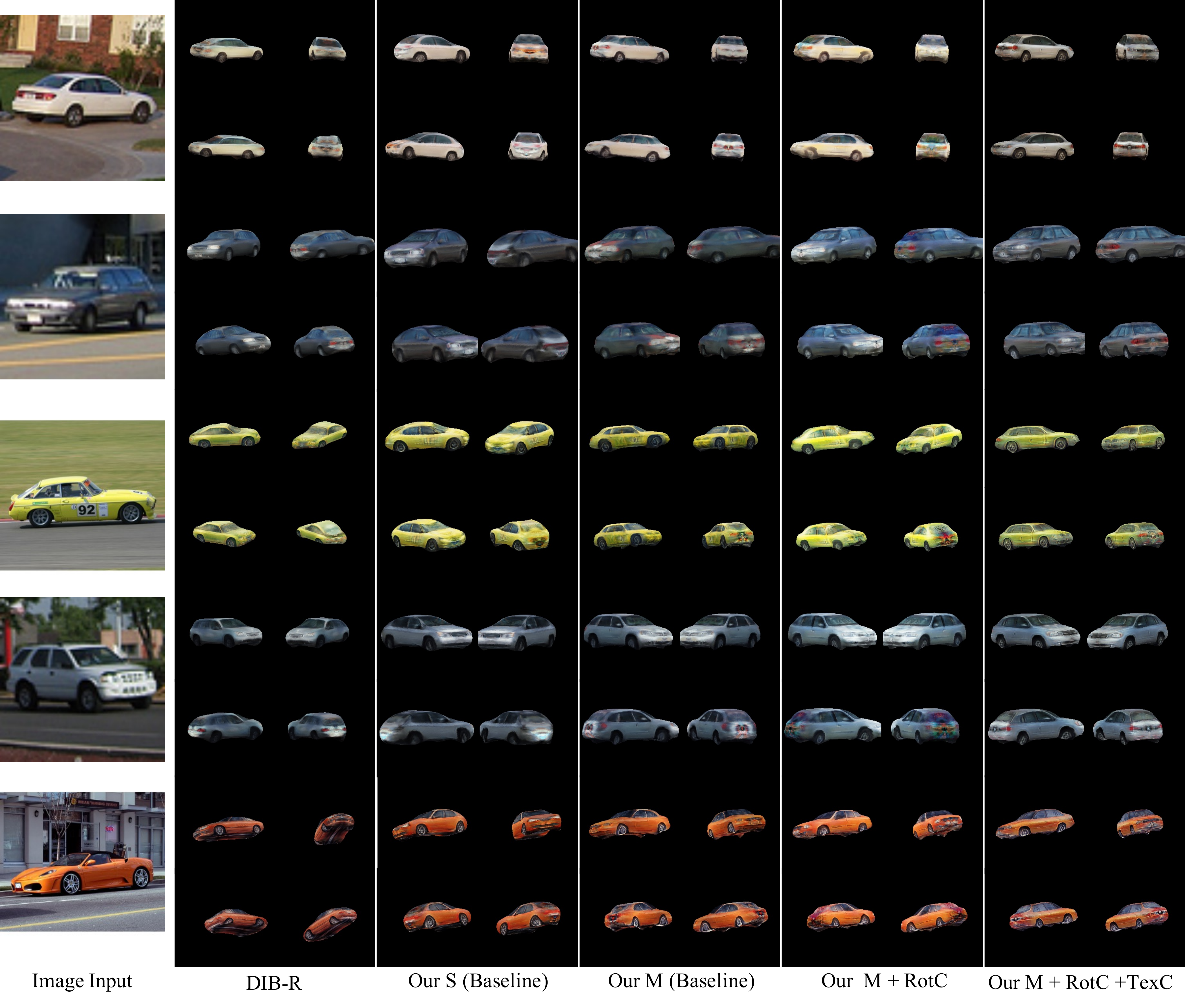}
  \vspace{-24pt}
  \caption{Textured 3D inference on Pascal3D+ cars \cite{xiang2014beyond}. We show renderings from 4 different views --   the original view and 3 other views by rotating the camera along the azimuth. We provide a comparison with DIB-R \cite{chen2019learning} and an ablation study showing how each of our components improves overall textured 3D synthesis. DIB-R (second column) suffers from view generalization -- the shapes are incoherent and textures are incomplete for novel views. Our single (S) template baseline(third column) using convolutional deformation improves overall shape but is not perfect. Using multiple (M) templates (fourth column), significantly improves the overall geometry of our synthesis. Rotation GAN consistency (RotC; fifth column) losses improve our shape further and also slightly improve texture. In the last column, adding our texture alignment consistency (TexC) loss, we get a strong improvement in the overall texture synthesis. Note that DIB-R uses ground truth (GT) camera for their renderings. The orange car in the bottom row has an erroneous GT camera and suffers from their view during final rendering. Our method reliably produces a correct estimate of the camera, fixing the erroneous GT camera estimations. 
  }
  \vspace{-9pt}
  \label{fig:car_qual}
 \end{figure*}

Our goal is to not only synthesize accurate 3D shapes but also realistic texture maps. However, since we train our network only with single-view images, the network predicts perfect texture and shape only for the same camera view, ignoring alignment of shapes and texture for novel views resulting in visible inconsistencies. We can view this as overfitting (biased) to the original view. Key to better novel view texture synthesis is avoiding such overfitting by building a robust prior over appearance, enabling sharing information between images to recover realistic texture for positions that are not visible from original view.
\begin{figure}[t]
  \centering
  \includegraphics[width=\linewidth]{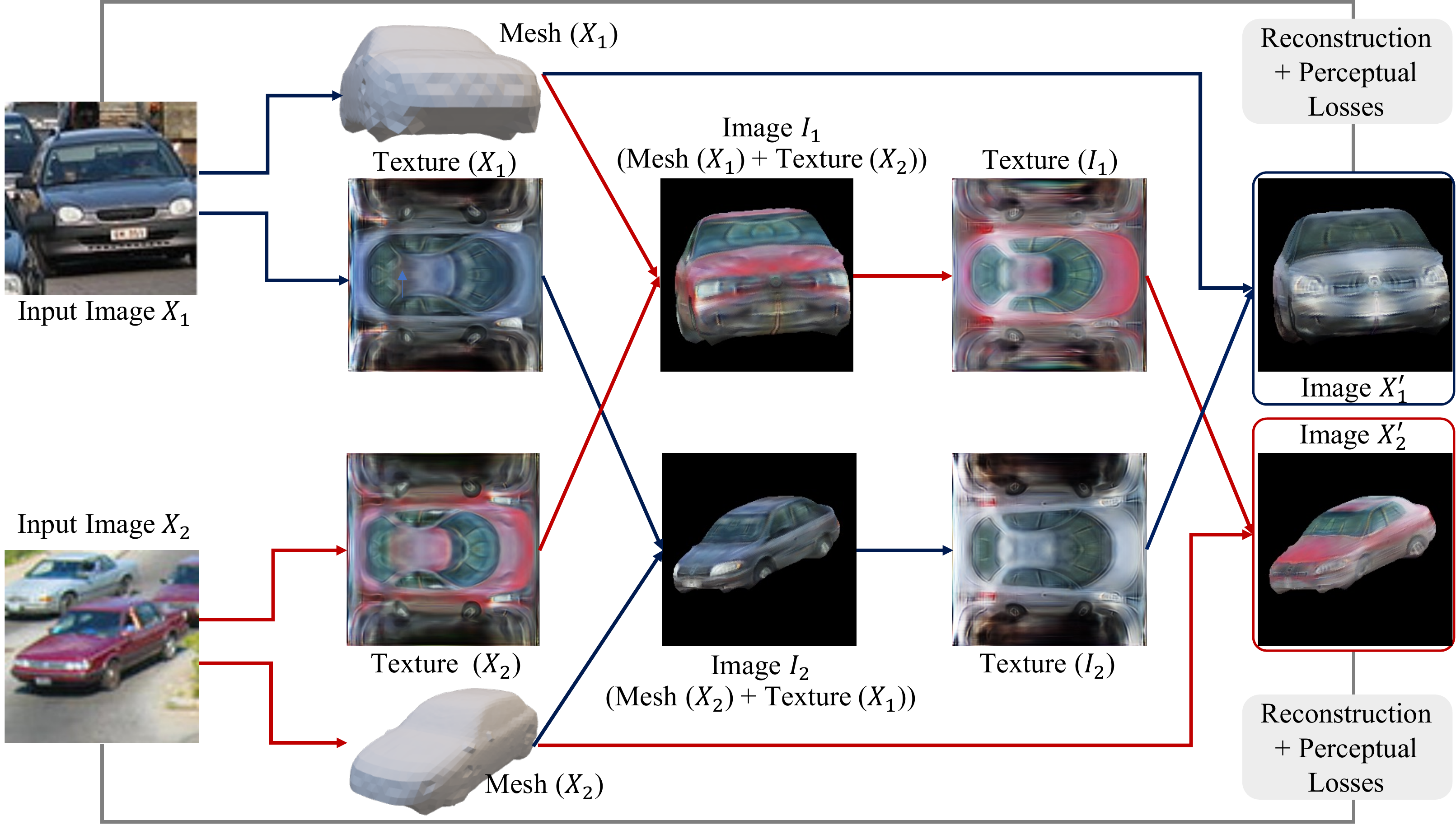}
  \vspace{-18pt}
  \caption{
 {\bf Texture Mesh Alignment Cycle Consistency} enables
texture representations to share when possible.  We
extract geometry and texture from randomly paired images
and render novel images with swapped textures between
these samples, then swap again and compare to original images.
This encourages the texture representation to store details like
(say) headlights in the same places, and so facilitates sharing textures
across models.
  }
    \vspace{-15pt}
  \label{fig:text_cyc}
 \end{figure}
We achieve this by texture mesh alignment cycle consistency (Fig.~\ref{fig:text_cyc}). Given two images, $X_1$ and $X_2$, and their corresponding camera poses $C_1$ and $C_2$, we infer their 3D mesh and textures. We then swap textures to reconstruct two new intermediate images:
 \vspace{-6pt}
 \begin{align*}
 \small
  & I_1 =  \text R(\text M(X_1), \text T(X_2), C_1) \\
  & I_2 =  \text R(\text M(X_2), \text T(X_1), C_2)
   \vspace{-15pt}
  \end{align*}
  \vspace{-18pt}\\
We now infer texture from these two new images and swap textures back again to reconstruct the original images:
  \vspace{-6pt}
 \begin{align*}
 \small
  & X_1^{'} =  \text R(\text M(X_1), \text T(I_2), C_1) \\
  & X_2^{'}  =  \text R(\text M(X_2), \text T(I_1), C_2)
   \vspace{-5pt}
  \end{align*}
    \vspace{-18pt}\\
The final renderings should match the input images since we swap textures back-and-forth.
The intuition behind this cycle constraint is to regularize the network from overspecializing to each example view (reduce bias). It also provides spatial cues for how the texture should behave to novel views and occluded areas by extracting representations from sampled images during back-and-forth swapping. This strongly helps in recovering accurate and realistic textures across novel views (see last column in Fig.~\ref{fig:car_qual}). %

\vspace{-10pt}
\section{Experiments}
In this section we describe our experimental evaluations.

{\noindent \bf{Datasets.}} We primarily use Pascal 3D+ dataset \cite{xiang2014beyond} - Car category for our evaluations.
We also evaluate our method for birds using CUBs \cite{welinder2010caltech} dataset.
We use same train-test split as that provided by DIB-R \cite{chen2019learning}. We use $7$ templates provided by Pascal3D+ dataset as our starting templates for cars after post-processing as described in our Section~\ref{sec:app}. For birds, we start with only a single template. 

{\noindent \bf Baselines and Ablations.} For our baselines, we mainly compare our method against the previous SOTA, DIB-R \cite{chen2019learning}.
We use their provided pretrained models for comparison. For DIB-R baseline, note that we used ground truth camera poses as they do not learn to infer camera poses. For all our evaluations, we use inferred camera poses. Another change from DIB-R is the use of reflection symmetry  for both predicting geometry and texture maps. DIB-R does not use symmetry for geometry as imposing this constraint hurts their overall quality of inferences. We also perform ablation studies to verify the extent of improvements over using multiple templates and single templates. The effectiveness of convolution deformation over linear layer deformation and also the role of our two cycle consistency losses.

{\noindent \bf Model Details.} We show our model pipeline in Fig.~\ref{fig:over_method}. We have two feature extractors: one from Resnet18 that is used to predict camera poses and template weights (when using over one template) and second for predicting deformation and texture maps. We predict camera poses and template weights using a single linear layer over Resnet18 features. Our predicted mesh shape then for an image is a simple weighted average over template vertices which we refine later using the deformation predicted. For predicting deformation and texture maps, we use a U-Net architecture. Our texture maps predicted are of $256\times256$ resolution. For our evaluations, we use deformation with $DOD = 4$ or $2\times2$ spatial resolution for cars and $DOD=256$ or $16\times16$ spatial resolution for birds. We sample deformation using a bi-linear grid interpolation for our predicted mesh by converting their world coordinates into UV coordinate using a spherical mapping procedure to get our final refined mesh. As we are not predicting deformation for each of the mesh vertices and only predict a single spatial deformation map, it allows us to sample deformation for many vertices without crippling computational efficiency.

For our GAN training, we tried two variants. For both these variants, we treated the rendered novel views as fake. In one variant, we used input training images as real and in the other rendered images from original view as real. We observed strong improvement over shapes when using input images as real and strong improvement in appearance when using rendered same view images as real. %

{\noindent \bf Evaluation.} As there is no ground-truth 3D geometry and texture maps to evaluate, we follow the same evaluation protocol used in DIB-R and evaluate the mask/silhouette projection accuracy. For view generalization, we report FID scores by rendering images from multiple views from the validation set and comparing them against the training data. Finally, we perform user study to evaluate human preferences comparing our method against the baselines. 

{\noindent \bf Results.} We provide quantitative and qualitative results in Fig.~\ref{fig:car_qual} and Table~\ref{tab:car}, respectively. From Fig.~\ref{fig:car_qual}, DIB-R outputs a good prediction for the original view and achieves a high 2D mIoU (see Table~\ref{tab:car}).   However, their texture and meshes do not generalize across views and results in poor FID  scores. Also, their camera is not learned and use ground truth camera for their rendering. Our S (single template) baseline is like DIB-R except the convolutional architecture for the mesh prediction. Our M (multiple templates) baseline improves the geometry considerably, however one can see unrealistic texture completions for novel views. Our M + RotC (Rotation GAN Cycle Consistency) improves the shape, makes the texture sharper and partially improves the texture predictions for novel views. Our final model with texture alignment consistency (TexC) improves our texture synthesis significantly. In the first row of Fig~\ref{fig:car_qual}, our final model outputs a realistic front view while others cannot. The realism of our synthesis is also evident from a strong improvement in FID scores (last two rows in Table~\ref{tab:car}.

\begin{figure}[t]
  \centering
  \includegraphics[width=\linewidth]{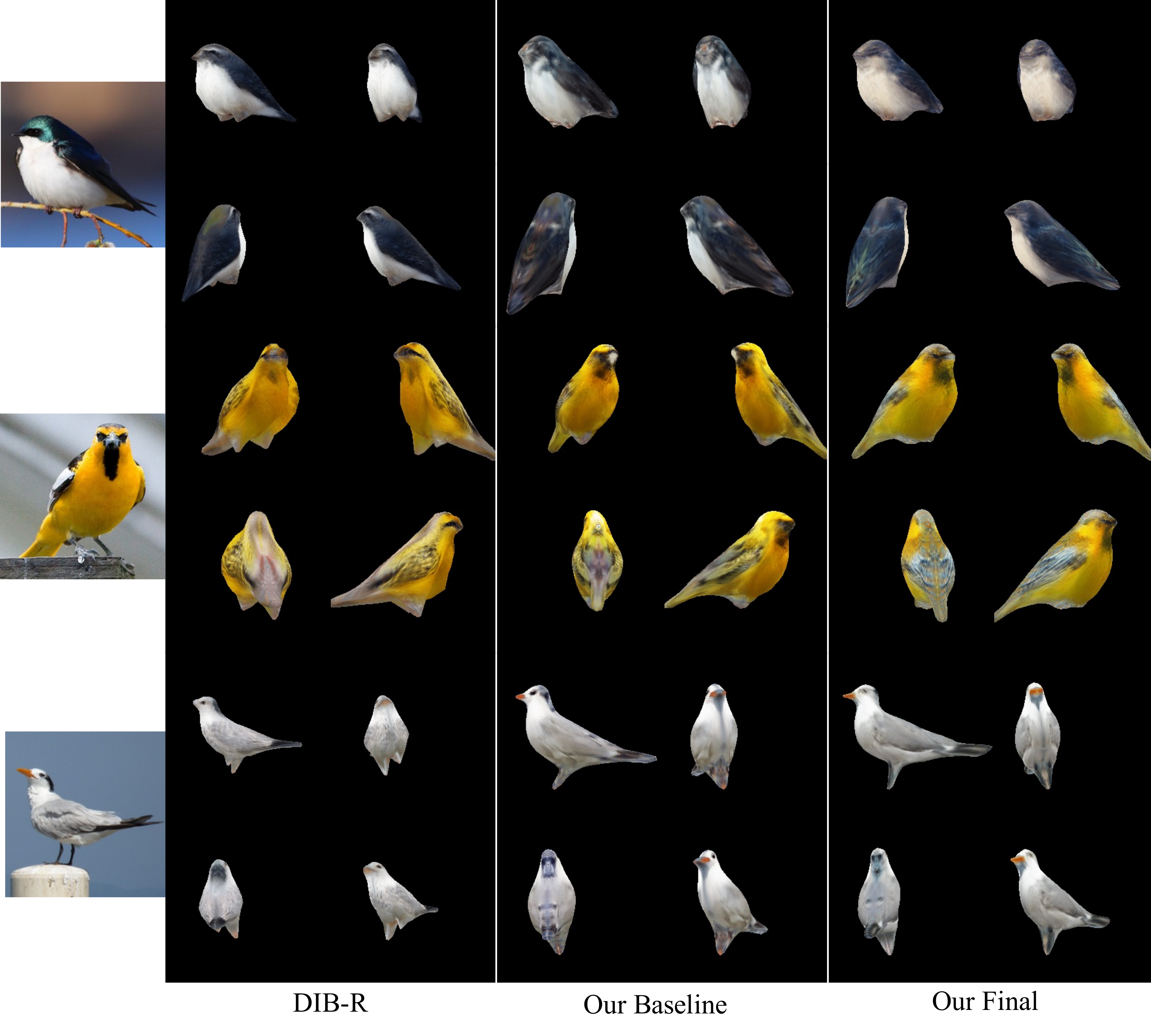}
  \vspace{-24pt}
  \caption{Textured 3D inference on the CUB dataset \cite{welinder2010caltech}. DIB-R renderings are consistent when viewed from original camera, but across novel views there are visible inconsistencies. Our baseline has improvements over shapes, but textures are sometimes not complete. Our proposed losses improve the shape further and produce realistic texture for occluded regions. For example, see feathers at the back for all rendered birds. Our method adds consistent texture for regions even if they are not visible from the original view.
  }
    \vspace{-10pt}
  \label{fig:birds_qual}
 \end{figure}

For evaluating view generalization, we also conducted a user study on the test set for cars on $220$ images. We provide users with input images and predictions rotated in $360^\circ$, along the azimuth, in GIF format so they can see the inconsistencies in the shape and texture easily. We ask users to select 3D models that have i) a more realistic texture, ii) a more realistic shape, and iii) overall a better 3D texture synthesis. Both our proposed losses and the convolutional framework significantly outperforms previous methods (see Table~\ref{table:user_study}). Users preferred our method as opposed to DIB-R $59.1\%$ of the time ($50\%$ is tie), and as opposed to our baseline without cycle consistency losses $53.54 \%$ of the time. The users also preferred our final model when we ask about texture and geometry specifically. We also observe improvement on birds as shown in Fig.~\ref{fig:birds_qual} and Table~\ref{table:bird}.

\begin{table}[t]
\vspace{-12pt}
\caption{Quantitative results on PASCAL3D+ Car dataset \cite{xiang2014beyond}. 2D mIOU accuracy predictions and GT, and FID scores between rendered images from various camera view-points and training images are reported. We also provide a full ablation study, showcasing improvements over each module.
DIB-R~\cite{chen2019learning} uses ground truth camera for their rendering during evaluation for mIOU (shown with *).}
\vspace{-9pt}
\resizebox{\linewidth}{!}
{
\begin{tabular}{lcccccc}
\toprule
Method & Temp.      & Deform & RotC & TexC & mIoU  & FID \\
\midrule
DIB-R \cite{chen2019learning} & S & Linear &     &      &   80.0*    &   198.0           \\
CMR \cite{kanazawa2018learning}   & S   & Linear &   &      & 64.0      &     --      \\
\midrule
Ours   & S & Conv   &     &      &  77.1    & 193.7           \\
Ours   & M  & Conv   &     &      &     76.9 &         175.5   \\
Ours   & M  & Conv   &  \cmark   &      &  {\bf 78.0}   &    171.6       \\
Ours   & M  & Conv   &     &  \cmark    &   77.3  &       166.8     \\
Ours   & M   & Conv   &   \cmark  & \cmark      &  {77.0}    &  {\bf157.6} \\
\bottomrule
\end{tabular}}
\label{tab:car}
\end{table}
\begin{table}[t]
\caption{Human preference evaluation on cars. We report user preferences to our synthesis using proposed convolutional deformation framework (Our Baseline) and additional proposed losses (Our Final) for three different studies. We ask users to choose the synthesis that had better texture, geometry and overall textured 3D synthesis.
}
\vspace{-9pt}
\centering
\resizebox{0.975\linewidth}{!}
{
\begin{tabular}{lccc}
\toprule
      & Texture  & Geometry & Overall \\
\midrule
DIBR vs Our Final & 54.1\% & 66.2\% & 59.1\% \\
Our Baseline vs  Our Final & 57.3\% & 59.6\% & 53.5\% \\
\bottomrule
\end{tabular}
}
\label{table:user_study}
\vspace{-5pt}
\end{table}

\begin{table}[t]
\vspace{-12pt}
\caption{Quantitative results on CUB dataset~\cite{welinder2010caltech}. 2D mIoU accuracy predictions and GT, and FID scores between rendered images from various camera viewpoints and training images are reported. For DIB-R~\cite{chen2019learning} and CMR~\cite{kanazawa2018learning}, we use mIOU reported in~\cite{chen2019learning} which uses a ground truth camera for their evaluation (shown with *) while we use a predicted camera for our mIOU evaluation.
}
\vspace{-9pt}
\resizebox{\linewidth}{!}
{
\begin{tabular}{lcccccc}
\toprule
Method & Temp.      & Deform & RotC & TexC & mIoU  & FID \\
\midrule
DIB-R~\cite{chen2019learning} & S & Linear &    &  &  75.7*    &         76.0          \\
CMR~\cite{kanazawa2018learning}   & S   & Linear &   &      & 73.8*      &     --      \\
\midrule
Ours   & S  & Conv   &     &      &     {62.5}  &         76.8 \\
Ours   & S  & Conv   &  \cmark   &      &  62.3  &     75.8     \\
Ours   & S   & Conv   &     & \cmark      &  \textbf{62.9}  & 75.0
 \\
Ours   & S   & Conv   &   \cmark  & \cmark      &  61.7 &  \textbf{70.3}\\   
\bottomrule
\end{tabular}
}
\label{table:bird}
\vspace{-5pt}
\end{table}

\vspace{-5pt}
\section{Conclusion}
In this paper, we introduced a new 3D inference pipeline, principally designed to display good view generalization. We use convolution deformation and show it has controllable and flexible deformation properties and can easily adapt to a particular choice of source image.  We also introduce two novel cycle consistency losses responsible for our high-fidelity textured 3D model synthesis from a single 2D image. Our losses enable the sharing of texture information and shape appearances for occluded regions. We show several improvements both qualitatively and quantitatively compared to the recent SOTA method. We hope our work will inspire future works to test their textured 3D inferences for view generalization and depart from 2D mIOU evaluations. Moreover, our overall framework can also apply to recently proposed unsupervised 3D inferences~\cite{goel2020shape, li2020self} and we believe these approaches would largely benefit from our design choices. Furthermore, because of consistent rendering across views, we believe our method would also be useful for rendering objects~\cite{bhattad2020cut} into scenes from novel views. 
\vspace{-13pt}
\section*{Acknowledgement}
\vspace{-1pt}
We thank Yuxuan Zhang and Wenzheng Chen  for providing us baseline code of~\cite{chen2019learning}. We also thank
David A. Forsyth for insightful discussions and his comments.

\appendix
\section{Implementation Details}

\textbf{Network architecture.} 
For predicting deformation and texture maps, we use a U-Net encoder-decoder architecture. The encoder is shared between the deformation and texture maps networks. The encoder contains 7 layers of Convolution-BatchNorm-LeakyReLU. Convolution layers have a kernel size of $5$, padding $2$, and stride $2$. At each layer, the number of filters doubles and goes as follows: $(32, 64, 128, 256, 512, 512, 512)$. The decoder for the texture network is a mirror-symmetric of the encoder and the number of filters are as follows $(1024, 1024, 1024, 512, 256, 128, 32)$. The number of filters in the decoder is two times of that encoder because the feature maps from the encoder skip to the corresponding decoder module and concatenates with the sequentially flowing feature maps. 
There are bilinear interpolation layers to upsample the feature maps at each layer.
For the deformation decoder, we follow a similar architecture but output from an earlier layer in the decoder. For example for $16\times16$ decoder map, there are 4 decoder layers with the following number of filters $(512, 512, 256, 128)$, and there are no skip connections from the encoder to the decoder. For both decoders, there is a final convolution layer that decreases the number of channels to $3$ for texture decoder they represent (R,G,B) channels and for the displacement decoder they represent the deformations (x,y,z) coordinates.

For predicting the camera parameters and template weights, we use a ResNet18 model pretrained on ImageNet. Same pretrained ResNet18 is also used by DIB-R for mesh prediction.
We use the pretrained convolutional layers of ResNet18 and stack a fully connected layer which outputs a feature vector with $200$ dimension. For camera parameters and template weights, the network branches out to two linear layers which decreases the number of feature vectors to the corresponding number of output parameters. 

We train our model on 8 GPUs with batch size of 16 per GPU, for 1200 epochs with learning rate of $1^{-4}$.

\section{Additional Ablation Studies}

\begin{table}[t]
\caption{Additional quantitative evaluations for measuring reconstruction accuracy from the original view.}
\centering
\begin{tabular}{lllll}
\toprule
& LPIPS $\downarrow$ & SSIM $\uparrow$ & PSNR $\uparrow$ & MSE $\downarrow$ \\
\midrule
DIB-R & 0.33 & 0.86 & 15.84 & 2023.7\\
Ours & \bf{0.31} & \bf{0.93} & \bf{16.18} & \bf{1888.1}  \\
\bottomrule
\end{tabular}
\label{table:quant_metric}
\end{table}

Table \ref{table:quant_metric} shows  quantitative evaluations for measuring reconstruction accuracy from the original view.
We compare ground-truth original view images with the rendered images in LPIPS \cite{zhang2018unreasonable}, SSIM, PSNR, and MSE.
Our results achieve better score than the baseline DIB-R on these metrics.

  \begin{figure}[t]
  \centering
  \includegraphics[width=\linewidth]{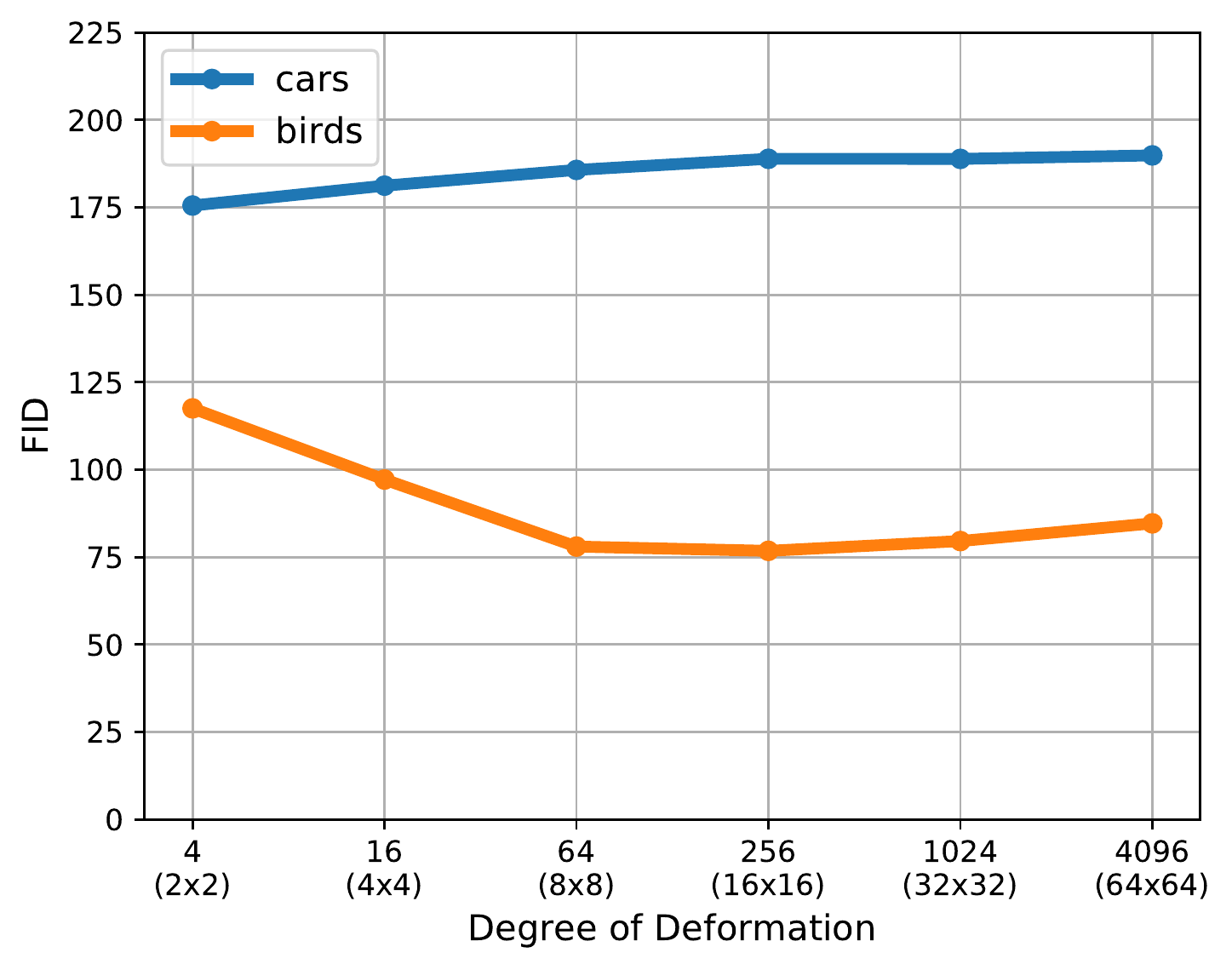}
  \caption{{\bf DOD Ablation study.} 
  For cars (a rigid object), increase in degree of deformation (DOD) results in poor generalization. Therefore, we use $DOD=4$. For birds (a non-rigid object), we need a model that is flexible to undergo a reasonable deformation. We find $DOD=256$ to generalize best across views. 
  }
  \label{fig:degree_of_deformation_ablation}
 \end{figure}

\begin{figure*}[t]
  \centering
  \includegraphics[width=0.98\linewidth]{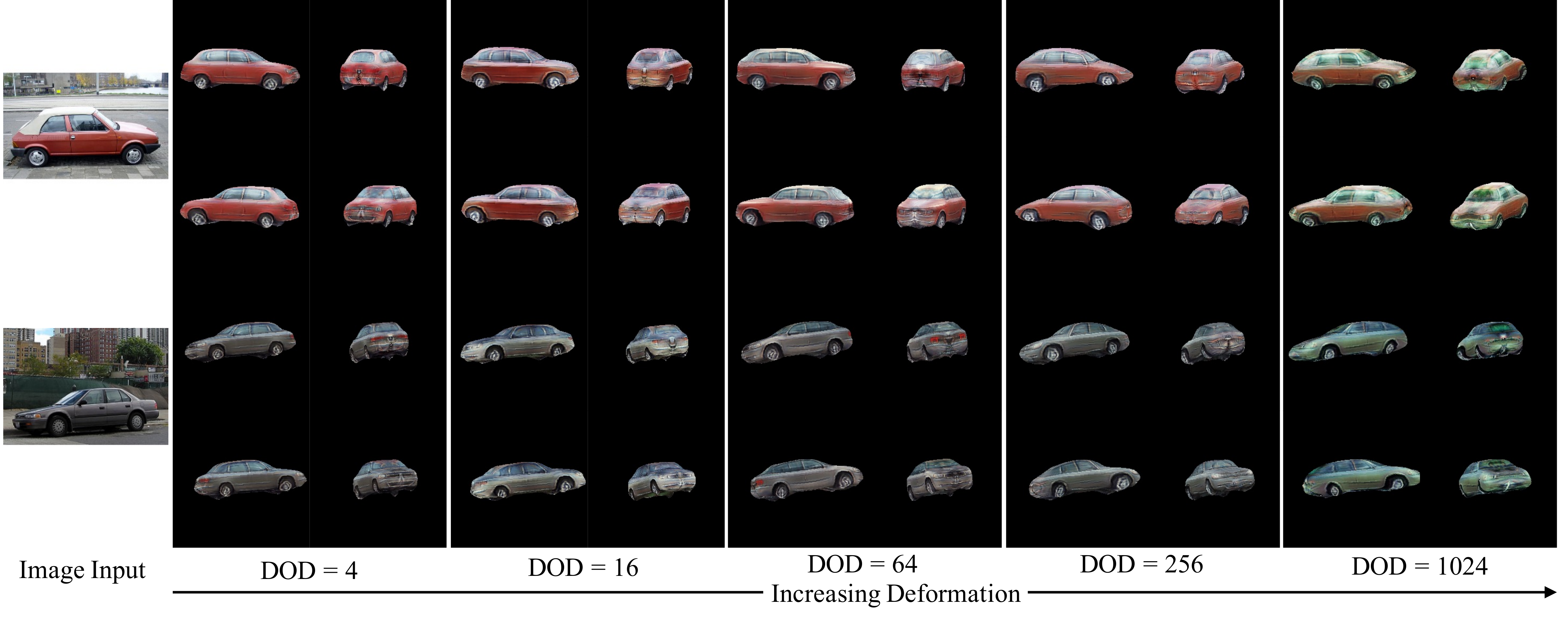}
  \caption{{\bf Degree of deformation ablation for cars.} 
  For cars (a rigid object), increasing the degree of deformation results in poor textured 3D model synthesis. Results degrade moving from left to right (with increasing deformation). 
  }
  \label{fig:degree_of_deformation_cars}
 \end{figure*}

 We provide an ablation study on the degree of deformation versus FID score for cars and birds categories in Fig. \ref{fig:degree_of_deformation_ablation}. The ablation study was conducted on our baseline model with convolutional deformation and with no consistency losses.
 For cars as rigid objects, we find increasing the degree of deformation results in poor generalization as measured in FID.
We use degree  of  deformation (DOD) of 4 in our experiments with cars. 
Our mesh templates also become useful in restricting the DOD to a such low value and generating realistic car 3D models.
Visual results of increasing degree of deformation for cars is provided in Fig. \ref{fig:degree_of_deformation_cars} and for birds in Fig. \ref{fig:degree_of_deformation}.
As shown in the figure,  increasing the degree of deformation results in poor textured 3D model synthesis. Results degrade moving from left to right (with increasing deformation) which is consistent with the FID metric.

For birds, as shown in Fig. \ref{fig:degree_of_deformation_ablation}, as a non-rigid object type, we find that we need a model that is flexible to undergo a reasonable deformation.  We find DOD of 256 to generalize best across views in our ablation study.
Additional freedom in the deformation also starts to hurt the performance.
 
\begin{figure}[t]
  \centering
  \includegraphics[width=\linewidth]{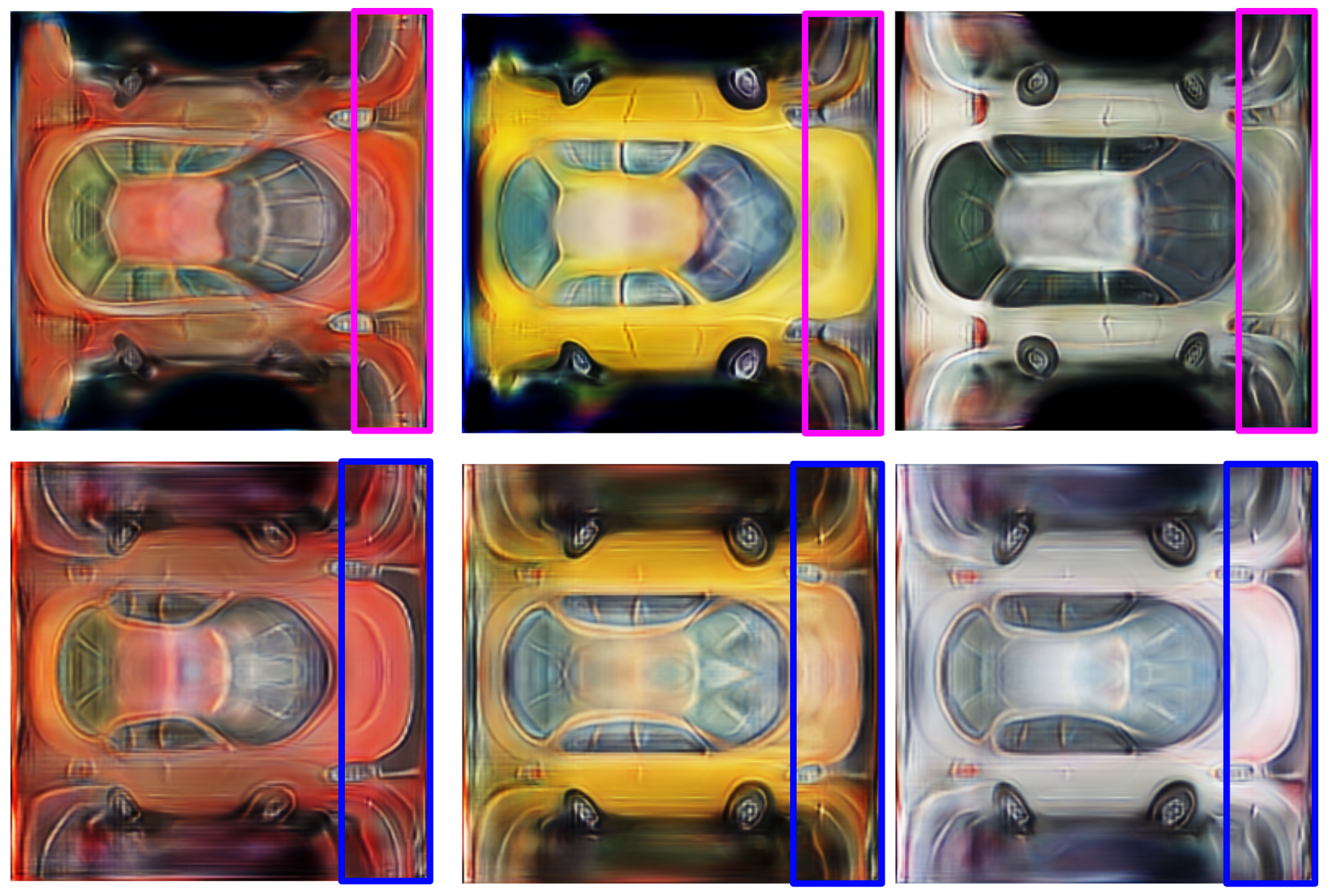}
  \caption{First row shows texture predictions from models trained without texture mesh alignment consistency and  second row with texture mesh alignment consistency. We provide purple and blue boxes for easy visualization to show the alignment.}
  \label{fig:text_alignment}
 \end{figure}

 One of our hypothesize is that the consistency losses we use in our model provides texture alignment across 3D models.
 This is not easy to test quantitatively. We provide texture predictions from models trained with and without texture mesh alignment consistency in Fig. \ref{fig:text_alignment}.
 In the figure, we find that the windshield in the first row starts at different places in the texture map whereas in the second row consistent placement of parts across examples is visible. Notice front headlights clearly aligned in the bottom row for all examples.
 
\section{Additional Results}

\begin{figure*}[t]
  \centering
  \includegraphics[width=\linewidth]{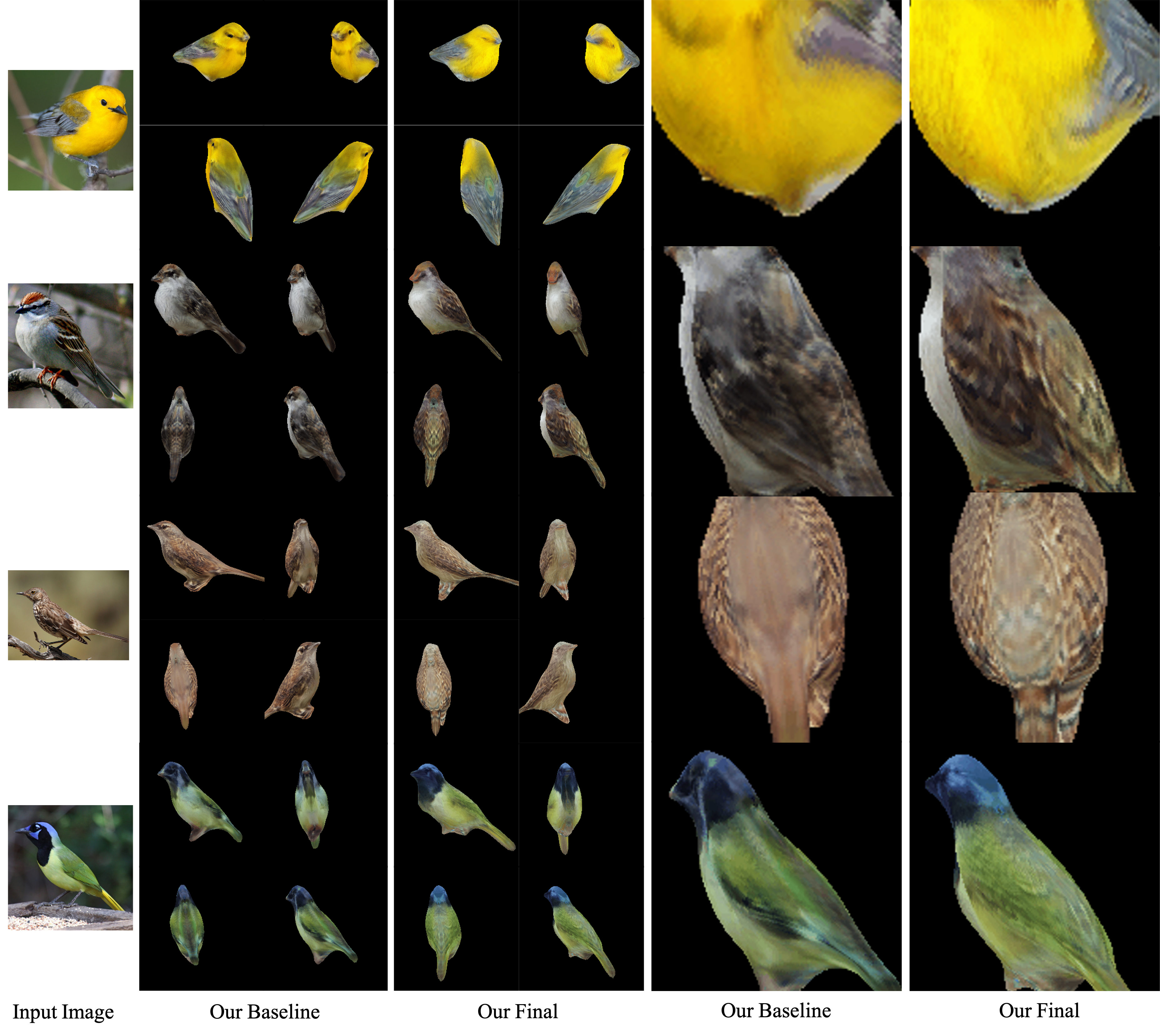}
  \vspace{-15pt}
  \caption{More example results on birds with $DOD=256$ or $16\times16$ resolution UV deformation map. Given input images on the left, we show results from four different views; the original view and three novel views for baseline and our model, respectively.
  Our baseline gets the overall color cues right but cannot add texture details, especially for occluded regions, and renders mostly with flat texture on the body. Our losses help to add detailed patterns in our final renderings. We show details in the last two columns to better visualize how our losses aids in improving overall texture. For example, for the yellow bird in the top row, our method adds detailed fur-like patterns on the body (last column) and our baseline falsely add's some black texture on the neck but it gets eyes better.
  }
  \label{fig:birds_supp}
 \end{figure*}

We provide more illustrative examples for birds in Fig. \ref{fig:birds_supp} showing overall improvements in shape and texture synthesis using proposed consistency losses. Our losses completes texture with appropriate patterns for occluded regions while our baseline creates a flat colored texture without details for occluded regions.
We show details in the last two columns to better visualize how our losses aids in improving overall texture for baseline and our results, respectively.  For the yellow bird, in the final rendering, our method adds detailed fur-like patterns on the body and our baseline falsely add’s some black texture on the neck.
Similar behaviours are observed for the other examples.

\section{User Study Set-up}
We evaluate our algorithm via a human subjective study. We perform pair-wise A/B tests deployed on the Amazon Mechanical Turk (MTurk) platform. We give users an input image, and two GIFs at once, each of which is synthesized from a different method.
We give users unlimited time to select which GIFs look more realistic. The left-right order and the image order are randomized to ensure fair comparisons.

Each test image, in total 220 of them, are compared 10 times, resulting in 2200 comparisons. Random chance results in $50\%$ preference.
In our studies, users pick our method when competed against our baseline and DIB-R for i) better texture, ii) better shape, iii) better overall synthesis as provided in the main paper. We found that the average time spent on each paired comparison was about 12 seconds.

\pagebreak
{\small
\bibliographystyle{ieee_fullname}
\bibliography{egbib}

\begin{thebibliography}{10}\itemsep=-1pt

\bibitem{bansal2018recycle}
Aayush Bansal, Shugao Ma, Deva Ramanan, and Yaser Sheikh.
\newblock Recycle-gan: Unsupervised video retargeting.
\newblock In {\em Proceedings of the European conference on computer vision
  (ECCV)}, pages 119--135, 2018.

\bibitem{bhattad2020cut}
Anand Bhattad and David~A Forsyth.
\newblock Cut-and-paste neural rendering.
\newblock {\em arXiv preprint arXiv:2010.05907}, 2020.

\bibitem{chang2015shapenet}
Angel~X Chang, Thomas Funkhouser, Leonidas Guibas, Pat Hanrahan, Qixing Huang,
  Zimo Li, Silvio Savarese, Manolis Savva, Shuran Song, Hao Su, et~al.
\newblock Shapenet: An information-rich 3d model repository.
\newblock {\em arXiv preprint arXiv:1512.03012}, 2015.

\bibitem{chen2019learning}
Wenzheng Chen, Huan Ling, Jun Gao, Edward Smith, Jaakko Lehtinen, Alec
  Jacobson, and Sanja Fidler.
\newblock Learning to predict 3d objects with an interpolation-based
  differentiable renderer.
\newblock In {\em Advances in Neural Information Processing Systems}, pages
  9609--9619, 2019.

\bibitem{chen2016infogan}
Xi Chen, Yan Duan, Rein Houthooft, John Schulman, Ilya Sutskever, and Pieter
  Abbeel.
\newblock Infogan: Interpretable representation learning by information
  maximizing generative adversarial nets.
\newblock In {\em Advances in neural information processing systems}, pages
  2172--2180, 2016.

\bibitem{choy20163d}
Christopher~B Choy, Danfei Xu, JunYoung Gwak, Kevin Chen, and Silvio Savarese.
\newblock 3d-r2n2: A unified approach for single and multi-view 3d object
  reconstruction.
\newblock In {\em European conference on computer vision}, pages 628--644.
  Springer, 2016.

\bibitem{dundar2020panoptic}
Aysegul Dundar, Karan Sapra, Guilin Liu, Andrew Tao, and Bryan Catanzaro.
\newblock Panoptic-based image synthesis.
\newblock In {\em Proceedings of the IEEE/CVF Conference on Computer Vision and
  Pattern Recognition}, pages 8070--8079, 2020.

\bibitem{dundar2020unsupervised}
Aysegul Dundar, Kevin Shih, Animesh Garg, Robert Pottorff, Andrew Tao, and
  Bryan Catanzaro.
\newblock Unsupervised disentanglement of pose, appearance and background from
  images and videos.
\newblock {\em IEEE Transactions on Pattern Analysis and Machine Intelligence},
  2021.

\bibitem{goel2020shape}
Shubham Goel, Angjoo Kanazawa, and Jitendra Malik.
\newblock Shape and viewpoint without keypoints.
\newblock {\em arXiv preprint arXiv:2007.10982}, 2020.

\bibitem{goodfellow2014generative}
Ian Goodfellow, Jean Pouget-Abadie, Mehdi Mirza, Bing Xu, David Warde-Farley,
  Sherjil Ozair, Aaron Courville, and Yoshua Bengio.
\newblock Generative adversarial nets.
\newblock In {\em Advances in neural information processing systems}, pages
  2672--2680, 2014.

\bibitem{henderson2020leveraging}
Paul Henderson, Vagia Tsiminaki, and Christoph~H Lampert.
\newblock Leveraging 2d data to learn textured 3d mesh generation.
\newblock In {\em Proceedings of the IEEE/CVF Conference on Computer Vision and
  Pattern Recognition}, pages 7498--7507, 2020.

\bibitem{heusel2017gans}
Martin Heusel, Hubert Ramsauer, Thomas Unterthiner, Bernhard Nessler, and Sepp
  Hochreiter.
\newblock Gans trained by a two time-scale update rule converge to a local nash
  equilibrium.
\newblock In {\em Advances in neural information processing systems}, pages
  6626--6637, 2017.

\bibitem{higgins2016beta}
Irina Higgins, Loic Matthey, Arka Pal, Christopher Burgess, Xavier Glorot,
  Matthew Botvinick, Shakir Mohamed, and Alexander Lerchner.
\newblock beta-vae: Learning basic visual concepts with a constrained
  variational framework.
\newblock In {\em International Conference on Learning Representations}, 2016.

\bibitem{isola2017image}
Phillip Isola, Jun-Yan Zhu, Tinghui Zhou, and Alexei~A Efros.
\newblock Image-to-image translation with conditional adversarial networks.
\newblock In {\em IEEE Conf. Comput. Vis. Pattern Recog.}, 2017.

\bibitem{jeon2020cross}
Subin Jeon, Seonghyeon Nam, Seoung~Wug Oh, and Seon~Joo Kim.
\newblock Cross-identity motion transfer for arbitrary objects through
  pose-attentive video reassembling.
\newblock {\em arXiv preprint arXiv:2007.08786}, 2020.

\bibitem{kanazawa2018learning}
Angjoo Kanazawa, Shubham Tulsiani, Alexei~A Efros, and Jitendra Malik.
\newblock Learning category-specific mesh reconstruction from image
  collections.
\newblock In {\em Proceedings of the European Conference on Computer Vision
  (ECCV)}, pages 371--386, 2018.

\bibitem{karacan2016learning}
Levent Karacan, Zeynep Akata, Aykut Erdem, and Erkut Erdem.
\newblock Learning to generate images of outdoor scenes from attributes and
  semantic layouts.
\newblock {\em arXiv preprint arXiv:1612.00215}, 2016.

\bibitem{karras2019style}
Tero Karras, Samuli Laine, and Timo Aila.
\newblock A style-based generator architecture for generative adversarial
  networks.
\newblock In {\em Proceedings of the IEEE conference on computer vision and
  pattern recognition}, pages 4401--4410, 2019.

\bibitem{karras2020analyzing}
Tero Karras, Samuli Laine, Miika Aittala, Janne Hellsten, Jaakko Lehtinen, and
  Timo Aila.
\newblock Analyzing and improving the image quality of stylegan.
\newblock In {\em Proceedings of the IEEE/CVF Conference on Computer Vision and
  Pattern Recognition}, pages 8110--8119, 2020.

\bibitem{kato2018neural}
Hiroharu Kato, Yoshitaka Ushiku, and Tatsuya Harada.
\newblock Neural 3d mesh renderer.
\newblock In {\em Proceedings of the IEEE Conference on Computer Vision and
  Pattern Recognition}, pages 3907--3916, 2018.

\bibitem{kulkarni2019csm}
Nilesh Kulkarni, Abhinav Gupta, and Shubham Tulsiani.
\newblock Canonical surface mapping via geometric cycle consistency.
\newblock In {\em International Conference on Computer Vision (ICCV)}, 2019.

\bibitem{li2020self}
Xueting Li, Sifei Liu, Kihwan Kim, Shalini De~Mello, Varun Jampani, Ming-Hsuan
  Yang, and Jan Kautz.
\newblock Self-supervised single-view 3d reconstruction via semantic
  consistency.
\newblock {\em arXiv preprint arXiv:2003.06473}, 2020.

\bibitem{liu2017unsupervised}
Ming-Yu Liu, Thomas Breuel, and Jan Kautz.
\newblock Unsupervised image-to-image translation networks.
\newblock In {\em Adv. Neural Inform. Process. Syst.}, 2017.

\bibitem{liu2019soft}
Shichen Liu, Tianye Li, Weikai Chen, and Hao Li.
\newblock Soft rasterizer: A differentiable renderer for image-based 3d
  reasoning.
\newblock In {\em Proceedings of the IEEE International Conference on Computer
  Vision}, pages 7708--7717, 2019.

\bibitem{liu2019deep}
Yu-Lun Liu, Yi-Tung Liao, Yen-Yu Lin, and Yung-Yu Chuang.
\newblock Deep video frame interpolation using cyclic frame generation.
\newblock In {\em AAAI}, 2019.

\bibitem{loper2014opendr}
Matthew~M Loper and Michael~J Black.
\newblock Opendr: An approximate differentiable renderer.
\newblock In {\em European Conference on Computer Vision}, pages 154--169.
  Springer, 2014.

\bibitem{lorenz2019unsupervised}
Dominik Lorenz, Leonard Bereska, Timo Milbich, and Bj{\"o}rn Ommer.
\newblock Unsupervised part-based disentangling of object shape and appearance.
\newblock In {\em CVPR}, 2019.

\bibitem{mardani2020neural}
Morteza Mardani, Guilin Liu, Aysegul Dundar, Shiqiu Liu, Andrew Tao, and Bryan
  Catanzaro.
\newblock Neural ffts for universal texture image synthesis.
\newblock {\em Advances in Neural Information Processing Systems}, 33, 2020.

\bibitem{meister2018unflow}
Simon Meister, Junhwa Hur, and Stefan Roth.
\newblock Unflow: Unsupervised learning of optical flow with a bidirectional
  census loss.
\newblock In {\em Thirty-Second AAAI Conference on Artificial Intelligence},
  2018.

\bibitem{nguyen2019hologan}
Thu Nguyen-Phuoc, Chuan Li, Lucas Theis, Christian Richardt, and Yong-Liang
  Yang.
\newblock Hologan: Unsupervised learning of 3d representations from natural
  images.
\newblock In {\em Proceedings of the IEEE International Conference on Computer
  Vision}, pages 7588--7597, 2019.

\bibitem{park2019semantic}
Taesung Park, Ming-Yu Liu, Ting-Chun Wang, and Jun-Yan Zhu.
\newblock Semantic image synthesis with spatially-adaptive normalization.
\newblock In {\em Proceedings of the IEEE Conference on Computer Vision and
  Pattern Recognition}, pages 2337--2346, 2019.

\bibitem{pavllo2020convolutional}
Dario Pavllo, Graham Spinks, Thomas Hofmann, Marie-Francine Moens, and Aurelien
  Lucchi.
\newblock Convolutional generation of textured 3d meshes.
\newblock {\em arXiv preprint arXiv:2006.07660}, 2020.

\bibitem{ravi2020accelerating}
Nikhila Ravi, Jeremy Reizenstein, David Novotny, Taylor Gordon, Wan-Yen Lo,
  Justin Johnson, and Georgia Gkioxari.
\newblock Accelerating 3d deep learning with pytorch3d.
\newblock {\em arXiv preprint arXiv:2007.08501}, 2020.

\bibitem{reda2019unsupervised}
Fitsum~A Reda, Deqing Sun, Aysegul Dundar, Mohammad Shoeybi, Guilin Liu,
  Kevin~J Shih, Andrew Tao, Jan Kautz, and Bryan Catanzaro.
\newblock Unsupervised video interpolation using cycle consistency.
\newblock In {\em Proceedings of the IEEE International Conference on Computer
  Vision}, pages 892--900, 2019.

\bibitem{schonfeld2020u}
Edgar Schonfeld, Bernt Schiele, and Anna Khoreva.
\newblock A u-net based discriminator for generative adversarial networks.
\newblock In {\em Proceedings of the IEEE/CVF Conference on Computer Vision and
  Pattern Recognition}, pages 8207--8216, 2020.

\bibitem{tulsiani2018multi}
Shubham Tulsiani, Alexei~A Efros, and Jitendra Malik.
\newblock Multi-view consistency as supervisory signal for learning shape and
  pose prediction.
\newblock In {\em Proceedings of the IEEE conference on computer vision and
  pattern recognition}, pages 2897--2905, 2018.

\bibitem{tulsiani2017multi}
Shubham Tulsiani, Tinghui Zhou, Alexei~A Efros, and Jitendra Malik.
\newblock Multi-view supervision for single-view reconstruction via
  differentiable ray consistency.
\newblock In {\em Proceedings of the IEEE conference on computer vision and
  pattern recognition}, pages 2626--2634, 2017.

\bibitem{wang2018pixel2mesh}
Nanyang Wang, Yinda Zhang, Zhuwen Li, Yanwei Fu, Wei Liu, and Yu-Gang Jiang.
\newblock Pixel2mesh: Generating 3d mesh models from single rgb images.
\newblock In {\em Proceedings of the European Conference on Computer Vision
  (ECCV)}, pages 52--67, 2018.

\bibitem{wang2019few}
Ting-Chun Wang, Ming-Yu Liu, Andrew Tao, Guilin Liu, Bryan Catanzaro, and Jan
  Kautz.
\newblock Few-shot video-to-video synthesis.
\newblock In {\em Advances in Neural Information Processing Systems}, pages
  5014--5025, 2019.

\bibitem{wang2018high}
Ting-Chun Wang, Ming-Yu Liu, Jun-Yan Zhu, Andrew Tao, Jan Kautz, and Bryan
  Catanzaro.
\newblock High-resolution image synthesis and semantic manipulation with
  conditional gans.
\newblock In {\em Proceedings of the IEEE conference on computer vision and
  pattern recognition}, pages 8798--8807, 2018.

\bibitem{welinder2010caltech}
P. Welinder, S. Branson, T. Mita, C. Wah, F. Schroff, S. Belongie, and P.
  Perona.
\newblock {Caltech-UCSD Birds 200}.
\newblock Technical Report CNS-TR-2010-001, California Institute of Technology,
  2010.

\bibitem{xiang2014beyond}
Yu Xiang, Roozbeh Mottaghi, and Silvio Savarese.
\newblock Beyond pascal: A benchmark for 3d object detection in the wild.
\newblock In {\em IEEE winter conference on applications of computer vision},
  pages 75--82. IEEE, 2014.

\bibitem{yu2021dual}
Ning Yu, Guilin Liu, Aysegul Dundar, Andrew Tao, Bryan Catanzaro, Larry Davis,
  and Mario Fritz.
\newblock Dual contrastive loss and attention for gans.
\newblock {\em arXiv preprint arXiv:2103.16748}, 2021.

\bibitem{zhang2019self}
Han Zhang, Ian Goodfellow, Dimitris Metaxas, and Augustus Odena.
\newblock Self-attention generative adversarial networks.
\newblock In {\em International Conference on Machine Learning}, pages
  7354--7363. PMLR, 2019.

\bibitem{zhang2018unreasonable}
Richard Zhang, Phillip Isola, Alexei~A Efros, Eli Shechtman, and Oliver Wang.
\newblock The unreasonable effectiveness of deep features as a perceptual
  metric.
\newblock In {\em Proceedings of the IEEE conference on computer vision and
  pattern recognition}, pages 586--595, 2018.

\bibitem{Zhao2018layout}
Bo Zhao, Lili Meng, Weidong Yin, and Leonid Sigal.
\newblock Image generation from layout.
\newblock In {\em IEEE Conf. Comput. Vis. Pattern Recog.}, 2019.

\bibitem{zheng2019joint}
Zhedong Zheng, Xiaodong Yang, Zhiding Yu, Liang Zheng, Yi Yang, and Jan Kautz.
\newblock Joint discriminative and generative learning for person
  re-identification.
\newblock In {\em Proceedings of the IEEE conference on computer vision and
  pattern recognition}, pages 2138--2147, 2019.

\bibitem{zhou2020rotate}
Hang Zhou, Jihao Liu, Ziwei Liu, Yu Liu, and Xiaogang Wang.
\newblock Rotate-and-render: Unsupervised photorealistic face rotation from
  single-view images.
\newblock In {\em Proceedings of the IEEE/CVF Conference on Computer Vision and
  Pattern Recognition}, pages 5911--5920, 2020.

\bibitem{zhou2020unsupervised}
Keyang Zhou, Bharat~Lal Bhatnagar, and Gerard Pons-Moll.
\newblock Unsupervised shape and pose disentanglement for 3d meshes.
\newblock {\em arXiv preprint arXiv:2007.11341}, 2020.

\bibitem{zhu2017unpaired}
Jun-Yan Zhu, Taesung Park, Phillip Isola, and Alexei~A Efros.
\newblock Unpaired image-to-image translation using cycle-consistent
  adversarial networks.
\newblock In {\em Proceedings of the IEEE international conference on computer
  vision}, pages 2223--2232, 2017.

\bibitem{zhu2018visual}
Jun-Yan Zhu, Zhoutong Zhang, Chengkai Zhang, Jiajun Wu, Antonio Torralba, Josh
  Tenenbaum, and Bill Freeman.
\newblock Visual object networks: Image generation with disentangled 3d
  representations.
\newblock {\em Advances in Neural Information Processing Systems}, 2018.

\bibitem{zhu2017face}
Xiangyu Zhu, Xiaoming Liu, Zhen Lei, and Stan~Z Li.
\newblock Face alignment in full pose range: A 3d total solution.
\newblock {\em IEEE transactions on pattern analysis and machine intelligence},
  41(1):78--92, 2017.

\end{thebibliography}
}

\end{document}